\documentclass{article}

\usepackage{arxiv}

\usepackage[utf8]{inputenc} 
\usepackage[T1]{fontenc}    
\usepackage{hyperref}       
\usepackage{url}            
\usepackage{booktabs}       
\usepackage{amsfonts}       
\usepackage{nicefrac}       
\usepackage{microtype}      
\usepackage{lipsum}
\usepackage{authblk}

\usepackage{float} 
\usepackage{times}
\usepackage{epsfig}
\usepackage{graphicx}
\usepackage{amsmath, amsfonts, amssymb}
\usepackage{bm}
\usepackage{algorithm}
\usepackage{algorithmicx}
\usepackage{comment}
\usepackage{fullpage}
\usepackage{tikz}
\usepackage{pifont}
\usepackage{amsthm}
\usepackage{subcaption}
\usepackage{amsmath,amsfonts,amssymb}
\usepackage{mathtools}
\usepackage{multirow}
\usepackage[noend]{algpseudocode}
\usepackage{booktabs}
\usepackage[export]{adjustbox}
\usepackage{url}

\usepackage[textsize=scriptsize,backgroundcolor=green]{todonotes}

\newcommand\numberthis{\addtocounter{equation}{1}\tag{\theequation}}

\usepackage{xspace}
\newcommand*{\eg}{\textit{e.g.}\@\xspace}
\newcommand*{\ie}{\textit{i.e.}\@\xspace}

\newcommand*{\vs}{\textit{vs.}\@\xspace}

\makeatletter
\newcommand*{\etc}{%
	\@ifnextchar{.}%
	{\textit{etc}}%
	{\textit{etc.}\@\xspace}%
}
\makeatother
\algtext*{EndWhile}
\algtext*{EndFor}
\algtext*{EndIf}
\algdef{SE}[DOWHILE]{Do}{doWhile}{\algorithmicdo}[1]{\algorithmicwhile\ #1}%

\makeatletter
\def\BState{\State\hskip-\ALG@thistlm}
\makeatother

\title{Extracting Optimal Solution Manifolds using Constrained Neural Optimization}

\author[1]{\textbf{Gurpreet Singh} \textsuperscript{\dag}}
\author[2]{\textbf{Soumyajit Gupta} \textsuperscript{\dag}}
\author[3]{\textbf{Matthew Lease}}
\affil[2]{Department of Computer Science}
\affil[3]{School of Information}
\affil[1]{The University of Texas at Austin}
\affil[ ]{\texttt{\{gurpreet, smjtgupta, ml\}@utexas.edu}}

\begin{document}

\maketitle

{\let\thefootnote\relax\footnote{{\dag contributed equally to this work.}}}

\begin{abstract}

Constrained Optimization solution algorithms are restricted to point based solutions. In practice, single or multiple objectives must be satisfied, wherein both the objective function and constraints can be non-convex resulting in multiple optimal solutions. Real world scenarios include intersecting surfaces as Implicit Functions, Hyperspectral Unmixing and Pareto Optimal fronts. Local or global convexification is a common workaround when faced with non-convex forms. However, such an approach is often restricted to a strict class of functions, deviation from which results in sub-optimal solution to the original problem. We present neural solutions for extracting optimal sets as approximate manifolds, where unmodified, non-convex objectives and constraints are defined as modeler guided, domain-informed $L_2$ loss function. This promotes interpretability since modelers can confirm the results against known analytical forms in their specific domains. We present synthetic and realistic cases to validate our approach and compare against known solvers for bench-marking in terms of accuracy and computational efficiency.

\end{abstract}

\noindent Constrained Optimization (CO) problems arise in a wide variety of applications such as Computational geometry \cite{gandomi2013bat}, Graphics \cite{christie2008camera}, Hyperspectral Imaging \cite{chen2011hyperspectral}, Information Retrieval \cite{metzler2007linear}, Economics \cite{yu2006scheduling}. Each of these fields construct a domain informed structure using an objective function, combined with a set of constraints to approximate a desired solution set. The bottleneck lies in efficiently searching the variable space to find this feasible solution set. The algorithmic difficulties in CO problems arise in the form of: limitations in converging to a global minimum in the presence of non-convex functions (objectives and constraints). In limited cases, under strict assumptions, convexified dual problems are achievable with bounded primal-dual solution gap. In this work, we attempt to address cases where convexification is either not achievable or the primal-dual solution gap cannot be bounded under practical considerations.

\begin{figure}[th]
    \centering
    \includegraphics[width=0.5\linewidth]{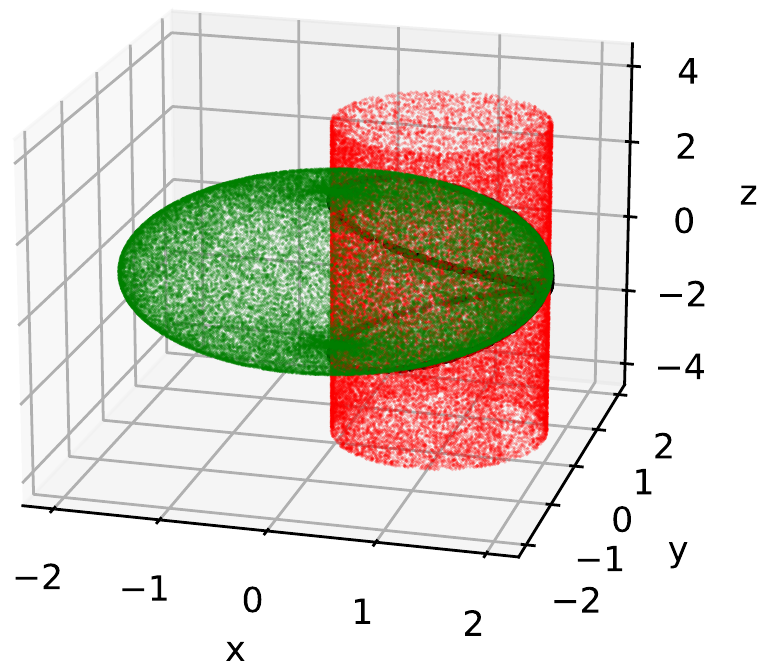}
    \caption{3D intersection manifold of a hollow sphere and a hollow cylinder identified using our constrained feed-forward neural optimization approach.}
    \label{fig:cyl_sph}
\end{figure}

 As in \cite{yen2006constraint}, we consider a simple example wherein the objective is to minimize $h(x) \equiv (f(x)-g(x))^2$ with $f(x)=x^2,g(x)=x+0.5$, such that $-2 \leq x \leq 2$. The optimal solution set in the absence of additional constraints is $x=\frac{1 \pm \sqrt{3}}{2}$ as shown in Fig. \ref{fig:nonlin} (left). Now consider an inequality constraint $-1 \leq x \leq 0$. Assuming a resolution of two decimal places in the finite (discrete) dimensional search space, the feasible solution set has size of $100$ among the $400$ possible candidates in the presence of this additional constraint, which is only $25\%$ of the domain. Under this setting an optimal value of $h(x)=0.00$ is achieved for $x=-0.37$. A similar precision limitation arises when the objective is to find the intersecting manifold of two point cloud objects. Our objective in this work is to extract this low-precision manifold, where the objective functions and constraints vary with applications. Under practical considerations, we are faced with the following challenges in a CO setting:
\begin{itemize}
    \item The objectives $f(x)$ and constraints $g(x)$ are non-convex, such that convexification imposes strict assumptions.
    \item An optimal solution manifold exists instead of a single global optimum, for \eg Pareto fronts. 
    \item True representation of $f(x),g(x)$ are unknown, or available as tuples $(x,f(x))$ and $(x,g(x))$ from point clouds.
\end{itemize}

Most existing CO solvers available in market as research and commercial tools are restricted to point based solutions with convexification assumptions. In a practical setting, any violation of these assumptions leads to sub-optimal solutions for the original problem. A point based solution scheme relies upon initializing a finite set of feasible points with a limited number converging to the optimal solution manifold. For \eg consider the solution scheme in \cite{gobbi2015analytical}, where a limited number of initialized points converge near the Pareto front, resulting in a rough description of the desired solution manifold. Our contributions in this work are as follows:
\begin{itemize}
\item We present constrained neural optimization solvers for both data and representation driven tasks.
\item Our framework can handle both single and multi-objective CO problems.
\item For a non-linear system of equations, our framework serves as a general purpose solver where the solution is in the form of a low precision indicator function defined continuously on the problem domain.
\item For graphics tasks, we can extract the implicit surfaces.
\item For multi-objective optimization, the network can approximate the Pareto frontier manifold.
\item For Linear Mixture model, it can approximate the end-members and abundances, starting from a random uniform initialization.
\end{itemize}

\section{Related Work}

A number of CO solution schemes have been developed for a range of problems. To list a few, (1) \textit{Penalty function} \cite{yeniay2005penalty} approaches where the constraint is enforced using a scaled penalization/barrier function in the resulting in an unconstrained optimization problem. (2) \textit{Preference of Feasible Solutions Over Infeasible Solutions} \cite{powell1993using} approaches ranks the feasible solutions with a higher fitness than infeasible ones. The rank-based selection scheme is based on the objective function and maps the feasible set into $(-\infty,1)$ rest to $(1,\infty)$. (3) \textit{Multi-objective Optimization} \cite{surry1995multi} approaches first ranks the solutions based on non-domination of constraint violations, followed by a re-ranking using the objective function. A factor based selection criteria is then used to accept/reject solutions. (4) \textit{Evolutionary algorithms} include Genetic Algorithms \cite{kumar1995genetic}, Simulated Annealing \cite{van1998constrained} and Particle Swarm Optimization \cite{zhang2005particle}, that relies upon a stochastic exploration of the solution space. (5) \textit{Divide and Conquer} \cite{shmoys1997cut} approaches leverage the form of the objective function to separate the solution space into convex subspaces, and consequently iterating between the unique minimizers of the sub-problems. 

Implicit surface are of special importance in the boundary evaluation of solid geometric models (CAD/CAM) and widely used in Computational Geometry and Computer Graphics. \cite{hartmann1998numerical} proposed numerical implicitization to treat intersection problems of both parametric and non-parametric surfaces. \cite{bajaj1988tracing} solves for the surface trace by calculating the implicit function values and its gradients at the intersection points.

For Linear Mixture Model problems, such as Hyperspectral Unmixing, either geometric or statistical approaches are used. Here, \textit{Pure Pixel} based algorithms include Pixel Purity Index (PPI) \cite{boardman1993automating}, N-FINDR \cite{winter1999n}, Vertex Component Analysis (VCA) \cite{nascimento2005vertex} and Successive Volume MAXimization (SVMAX) \cite{chan2011simplex}. These approaches assume the presence of at least one pure pixel per end-member in the data. \textit{Minimum Volume} based algorithms include Minimum Volume Simplex Analysis (MVSA) \cite{li2008minimum}, SISAL \cite{bioucas2009variable}, Minimum Volume Enclosing Simplex (MVES) \cite{chan2009convex} and VolMin \cite{fu2016robust}. Here an additional term is employed in the objective function, to minimize the volume of the simplex formed by the end-members. Recently, a number deep auto-encoder networks have been proposed:  End-net \cite{ozkan2018endnet}, Auto-Enc \cite{palsson2018hyperspectral} DCAE and \cite{khajehrayeni2020hyperspectral}, where the encoder is expected to capture the end-members in its latent representation, while the decoder attempts to minimize the loss in reconstructing the input spectrum.

For multi-objective optimization problems, Pareto optimality describes a trade-off scenario such that the solution is not biased towards any of the objectives. In other words, a gain in one objective should not result in the loss of even one other objective. The Pareto front is defined as the set of all such Pareto optimal points. \cite{gobbi2015analytical} present analytical solutions for a number of functional forms as case studies, and comparison against a numerical solution obtained using Genetic Algorithm. Similarly, \cite{ghane2015new} present a comparison of scalarization approaches for another set of multi-objective problems and benchmarks them against known analytical solutions.

\begin{figure*}[t]
    \centering
    \includegraphics[width=0.8\linewidth]{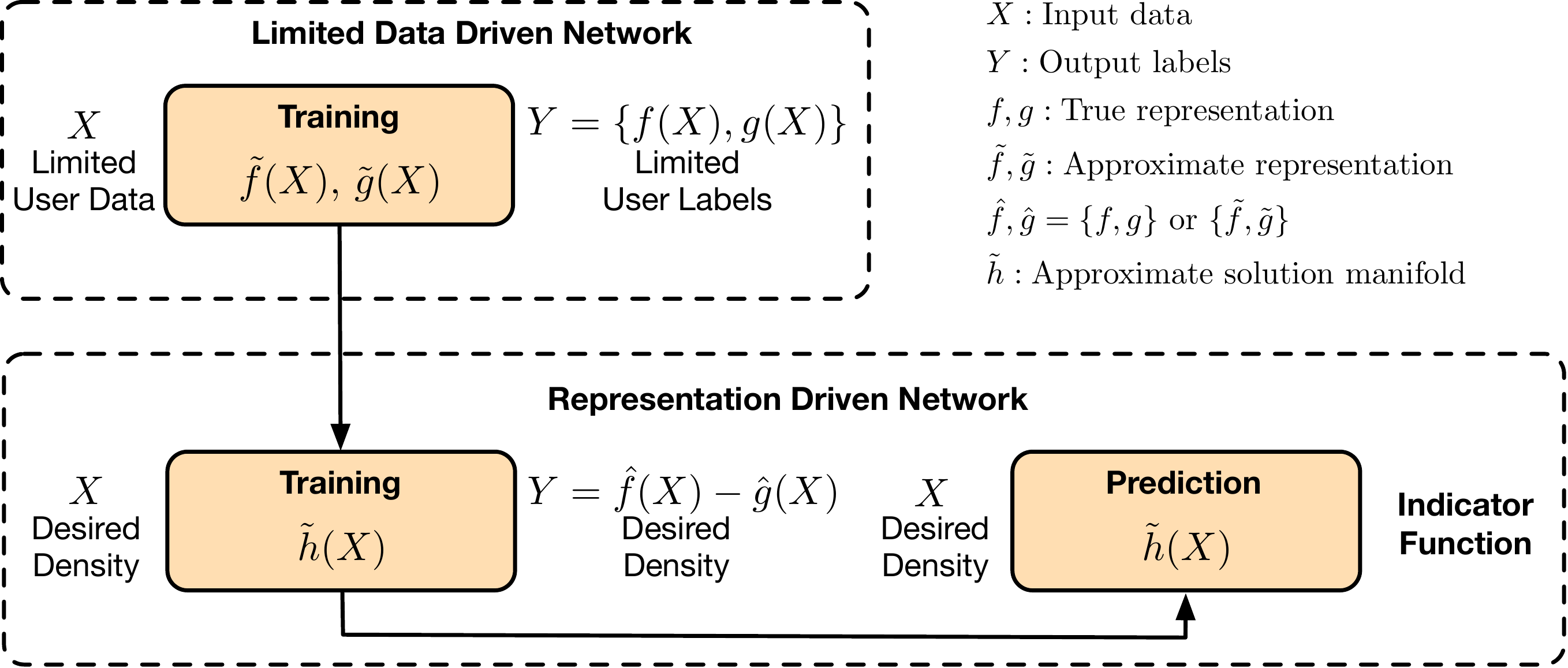}
    \caption{Overview of our constrained neural optimization framework. The framework allows labels specified either as discrete data points (data driven) or continuous functions (representation driven) of input data. In either case, a domain informed loss is used to approximate the true intersection manifold $h(X)=f
    (X)-g(X)$ as $\tilde{h}(X)$. The indicator function then describes a lower dimensional intersection manifold corresponding to $h(X^*)=0$, where $X^*$ is the set of optimal points in this setting.}
    \label{fig:arch}
\end{figure*}

\section{Formulation}
We consider the following general optimization problem with objective functions $f_i(x)$ along with  equality and inequality constraints defined as:
\begin{align*}
    &\underset{x}{min} \, f_{i}(x), \\
    &\text{s.t.} \quad p_j(x)=0, \text{ and } \quad q_k(x)\leq 0
\end{align*}
Here, $1 \leq i,j,k \in \mathbb{I}$, $x \in \mathbb{R}^n$ with $n \in \mathbb{Z}_+$ is a positive integer and $f_i,p_j,q_k:\mathbb{R}^n \rightarrow \mathbb{R}$ are continuous and differentiable functions. In what follows, we present a number of applications where the form and number of objectives and constraints vary subject to the problem at hand. 

\section{Network Design}

Our network design consists of dense feed forward neural networks, with width $(w)$ and depth $(d)$ representing the number of neurons in each internal layer and the number of internal layers respectively. The networks are low-weights (compact), and do not require use of Dropout, Batch Normalization or kernel regularization. The $(w,d)$ corresponding to the problem setting are reported in the result section.

\subsection{Data Driven Network}

In almost all practical problems, we parameterize discrete observations as models to closely represent an unknown absolute truth function $f$. Neural networks generally allow us to construct an approximate representation of $f$. The Limited Data Driven block of our framework in Fig. \ref{fig:arch} can approximate the target functions $f,g$ as $\tilde{f},\tilde{g}$, given training data $(X,Y)$. . This portion of the network can be trained in parallel for all available forms of $f,g$. A combination of these network instances are then used in the Representation Driven block as training pairs.

\subsection{Representation Driven Network} 

A problem specific loss is prescribed to approximate the true manifold $h$ as $\tilde{h}$. Although in Fig. \ref{fig:arch}, the description of $h$ as $h=f-g$ corresponds to finding an intersection manifold, we later show in the results section that by changing the form of $h$ we can solve an array of CO problems. Furthermore, if a well accepted representation of $f,g$ is already known, the Limited Data block can be skipped altogether.

\subsection{Loss Informed Manifold Detection}

The simplicity of our framework lies in the explicit prescription of a loss function corresponding to a user specified target. In a practical setting with limited data, a CO problem is a natural choice. The indicator function learned by Representation Driven network describes the lower dimensional solution manifold corresponding to $h(X^*)= 0$, where $X^*$ is the set of optimal points for all of our problem settings.

\section{Results}

\subsection{Setup and Training}

All experiments were done on a setup with Nvidia 2060 RTX Super 8GB GPU, Intel Core i7-9700F 3.0GHz 8-core CPU and 16GB DDR4 memory. We use the Keras \cite{chollet2015} library running on a Tensorflow 1.14 backend with Python 3.7 to train the networks in this paper. For optimization, we use AdaMax \cite{kingma2014adam} with parameters (\textit{lr}= 0.001) and $2000$ steps per epoch and $10$ epochs unless otherwise specified.

\subsection{Case I: Solving a Non-Linear System of Equations}

\begin{figure}[h]
    \centering
   	\includegraphics[width=0.4\linewidth]{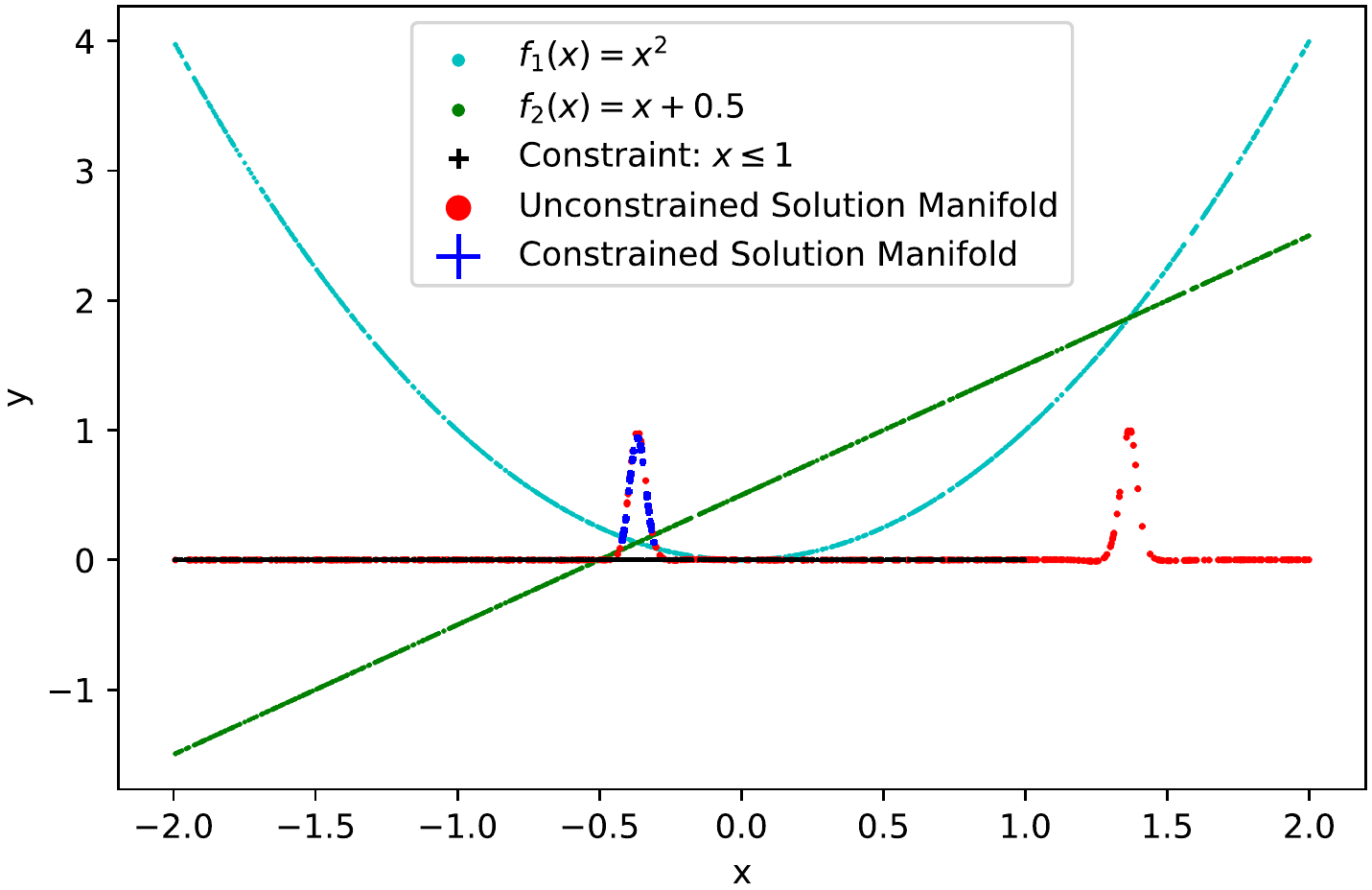}
   	\includegraphics[width=0.4\linewidth]{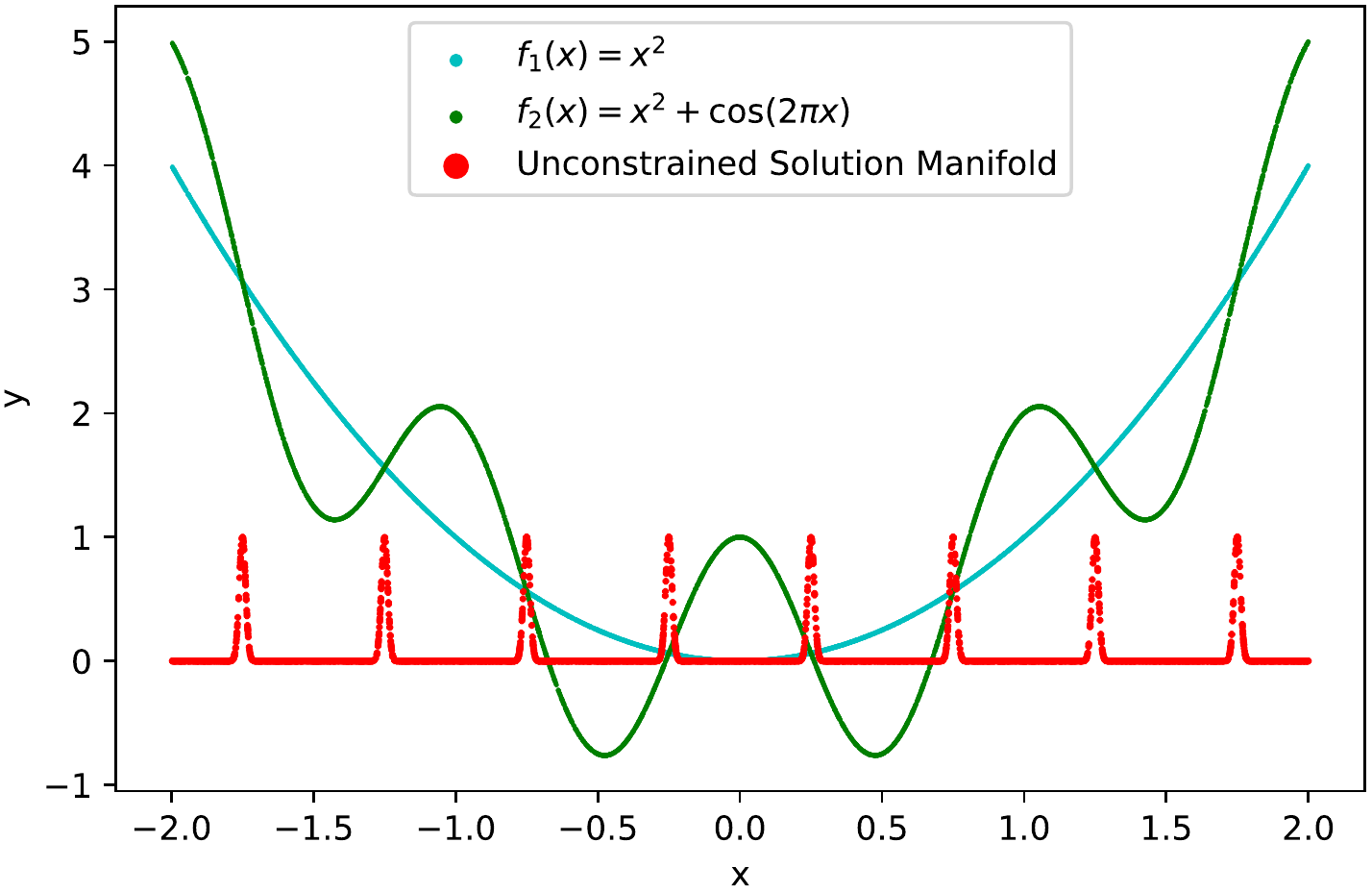}
    \caption{Intersection of non-linear curves: parabola and line (left) and parabola and parabolic cosine (righ)}
    \label{fig:nonlin}
\end{figure}

We first present two simple examples where the objective is to find all solutions to a system of non-linear equations in finite domain as shown in Fig. \ref{fig:nonlin}. For both of these problems the network size is fixed at $(w,d)=(4,4)$. The objective is to construct an indicator function ($I: R \rightarrow R$) that approximately indicates the location of intersections while approximating the manifold $h(x) = f_1(x) - f_2(x)$ such that $h(x^{*}) = 0$ where $x^{*}$ is the solution set.
Since the problem is formulated as a minimization problem, enforcing a constraint $x\leq 1$ results in only one solution as shown in Fig. \ref{fig:nonlin} (left), manually calculated in the introduction section.

\subsection{Case II: Implicit Surfaces}
\begin{figure}[h]
    \centering
	\includegraphics[width=0.4\linewidth]{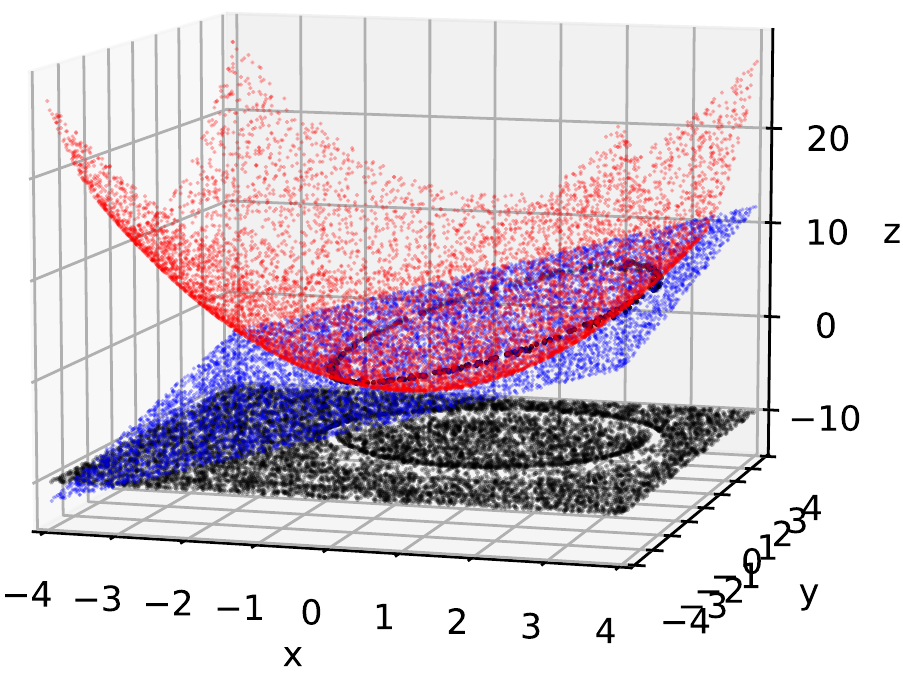}
	\includegraphics[width=0.4\linewidth]{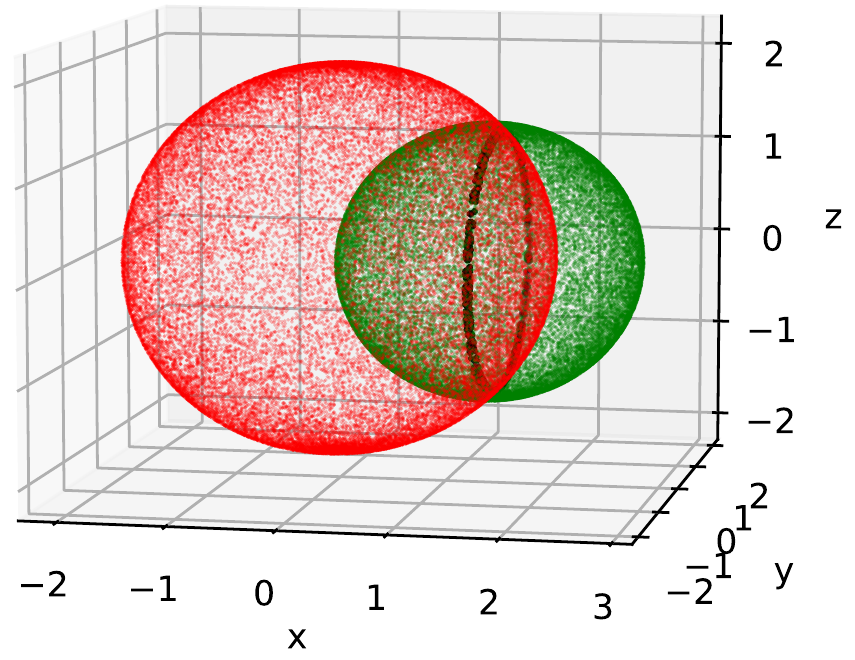}
    \caption{Intersection of 2D and 3D surfaces: plane and parabola (left), and two hollow spheres (Right). Notice the solution manifold exists over the entire domain, but only peaks along the appropriate intersection boundary for the 2D surfaces. A similar plot for 3D surfaces is not visually viable and therefore the solution set on the manifold is shown.}
    \label{fig:impli}
\end{figure}

In this next set of examples we ramp our framework to approximate implicit surfaces resulting from intersection of higher dimensional manifolds. Figs. \ref{fig:impli}, show the approximate implicit surfaces (color black) for visual ratification. Notice that the indicator function peaks near the intersecting region in Fig. \ref{fig:impli} (left). The network design is held at $(w,d)=(8,4)$.
Fig. \ref{fig:cyl_sph} shows a sphere of radius 2 centered at (0,0,0) with the z-axis aligned cylinder of radius 1 centered at (1,0,0). Fig. \ref{fig:impli} (right) shows two spheres centered at (0,0,0) and (1,0,0), with radii 2 and 1, respectively. These examples serve as test cases to aid network interpretability for higher dimensional manifold extraction problems in latter experiments on HSI unmixing problem. 

\subsection{Case III: End-Member Extraction}
In a Linear Mixture Model (LMM), each image pixel $y \in \mathbb{R}^F$ represents a data point and is a linear combination of a set of $K$ end-members $A=[a_1,\ldots,a_K] \in \mathbb{R}_+^{F \times K}$, with different fractions $b=[b_1,\ldots,b_K] \in \mathbb{R}_+^{K}$. In addition, $\gamma$ is an additive noise (assumed zero mean Gaussian) to account for sensor noise and illumination variation \etc. The LMM can thus be described as:
\begin{align} 
y = \sum_{k=1}^{K}a_kb_k + \gamma, \quad \text{s.t.} \, a_k,b_k \geq 0, \, \sum_{k=1}^Kb_k=1 
\label{eq:mix}
\end{align}
LMM on remote sensing Hyperspectral Images (HSI) assumes the entries of $A,b$ are positive, since spectral signatures are all positive, end-members contribute additively, and their contributions sum up to one. Assuming that the data is projected onto the signal subspace $\mathcal{S}$, of dimension $K$ and the end-members $[a_1,\ldots,a_K]$ are affinely independent (\ie $a_k-a_1$ for $k=2,\ldots,K$ are linearly independent), the simplex cone is defined as conv(A). Since we are interested in the minimum volume approaches, we seek to minimize the the volume of this convex cone to arrive at the solution.
\begin{align}
    V(A) = \frac{|det(A)|}{K!}
\end{align}
The equivalent problem then becomes:
\begin{align}
    \underset{A,B}{min} \, \|Y-AB\|_F^2 + \lambda det(A^TA)
\end{align}
Here $\lambda$ is a hyper-parameter that balances the trade-off between the regression loss and the minimum volume loss. We select the Samson \cite{zhu2014spectral} and Jasper \cite{zhu2014structured} HSI datasets for comparison and benchmarking. Samson contains $9025$ pixels, $156$ features and $3$ end-members, while Jasper has $10000$ pixels, $200$ features and $4$ end-members. The comparative results show Mean Square Error (MSE) and Spectral Angle Distance (SAD) for recovery of $A$.
\begin{align*}
    MSE &= \frac{1}{N} \sum \limits_{i=1}^N (a_i - \hat{a}_i)^2 \\
    SAD &= arccos\left(\frac{1}{K} \sum \limits_{j=1}^K \frac{a_j^T\hat{a}_j}{\|a_j\|\|\hat{a}_j\|}\right)
\end{align*}

The key advantages of our approach compared to convexified and deep end-member extraction techniques are:
\begin{itemize}
    \item Our network can approximate the true solution with random initialization, unlike approaches like Volmin \cite{fu2016robust} that requires explicit initialization from algorithms like SISAL \cite{bioucas2009variable}. 
    \item Our low-weights network attains lower error values faster than auto-encoder based deep algorithms.
    \item Unlike ADMM based convex updates, where changes in $B$ in the current iteration can affect changes in $A$ in the next iteration, our network treats $A$ as an invariant structure to be approximated, irrespective of $B$.
    \item Compared to other Deep approaches, since our network is significantly low-weight and does not use redundancy removal techniques like Dropout or Batch Normalization.
    \item The number of our network parameters is explicitly equal to $2\times K \times F$, irrespective of the image size.
\end{itemize}

\begin{figure}[h]
    \centering
	\includegraphics[width=0.8\linewidth]{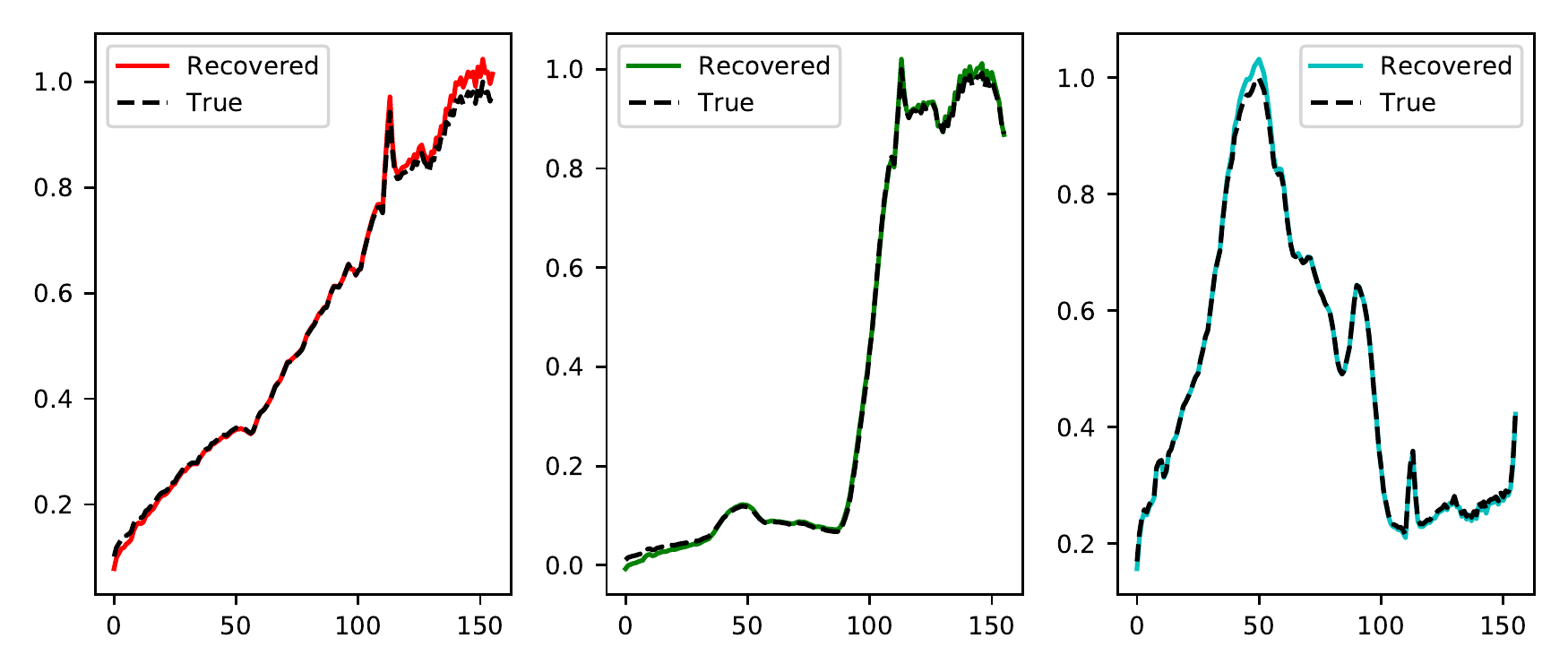}
    \caption{Plot of True \vs Recovered Spectra for Samson. X-axis represents bands, and y-axis represents Reflectance.}
    \label{fig:samson_spec}
\end{figure}
\begin{figure}[h]
    \centering
	\includegraphics[width=0.7\linewidth]{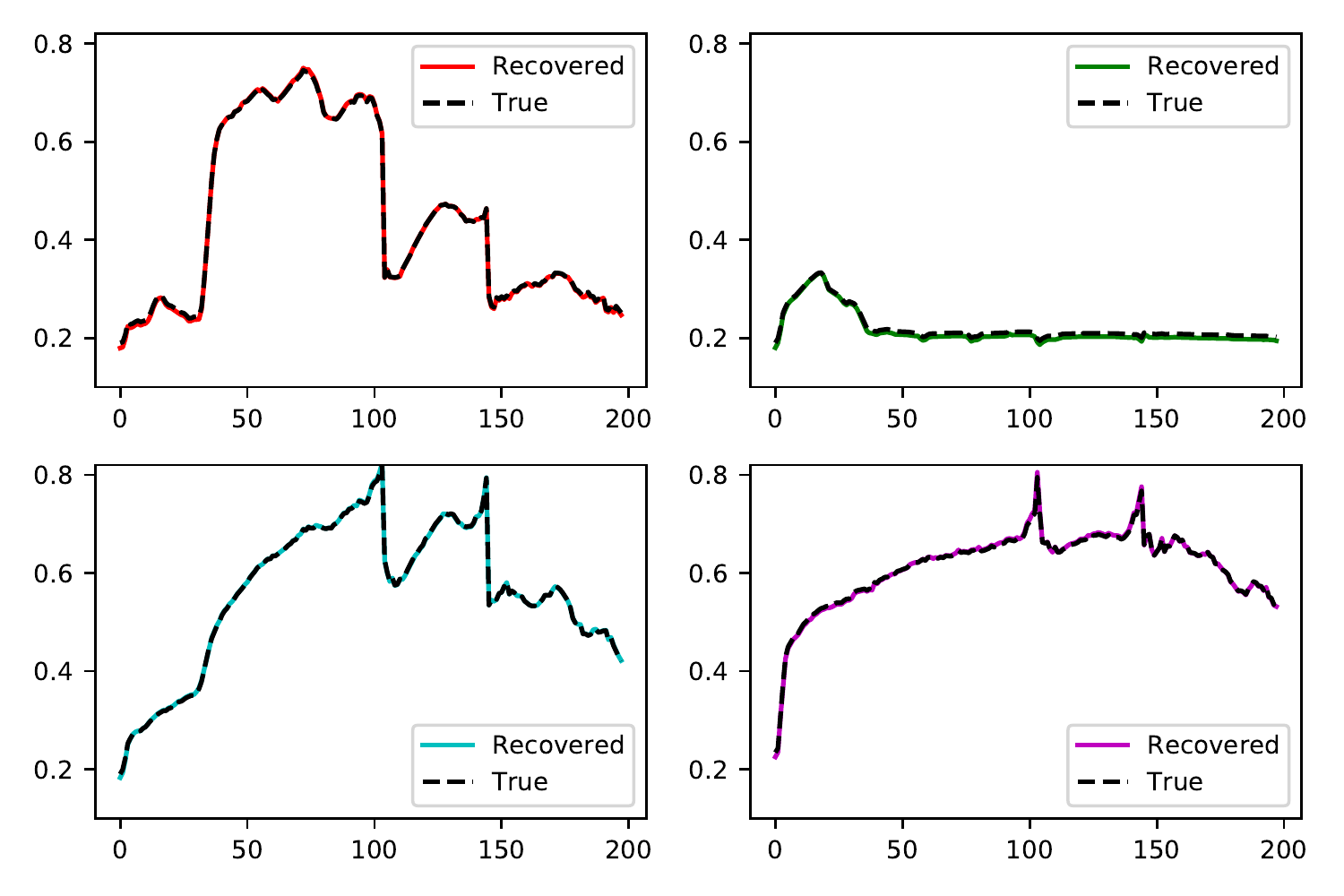}
    \caption{Plot of True \vs Recovered Spectra for Jasper. X-axis represents bands, and y-axis represents Reflectance.}
    \label{fig:jasper_spec}
\end{figure}
\begin{figure}[h]
    \centering
    \includegraphics[trim={0 0 0 0},clip, width=0.6\linewidth]{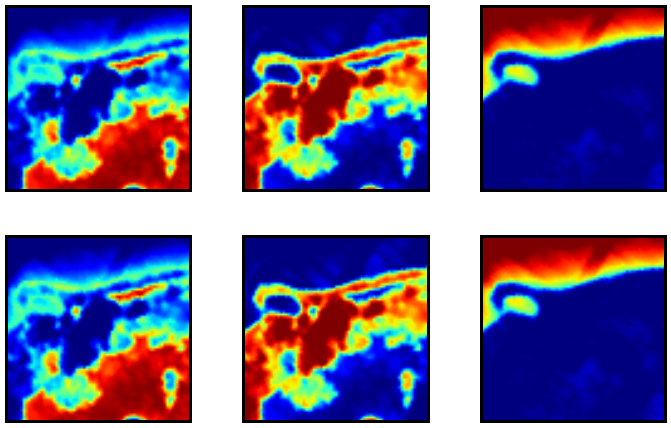}
    \caption{Top: Recovered. Bottom: True Abundance Maps for Samson. The colormap is on scale from 0 (blue) to 1 (red).}
    \label{fig:samson_abun}
\end{figure}

Fig. \ref{fig:samson_spec} shows the plot of the true and recovered spectrum for the 3 end-members for Samson dataset. Notice that the recovered spectra are not missing any spectral features when compared to the true end-members. The minor differences are just baseline variations in the spectra. Fig. \ref{fig:samson_abun} shows the corresponding abundance maps. 

\begin{figure}[h]
    \centering
    \includegraphics[trim={0 1cm 0 1cm},clip, width=0.8\linewidth]{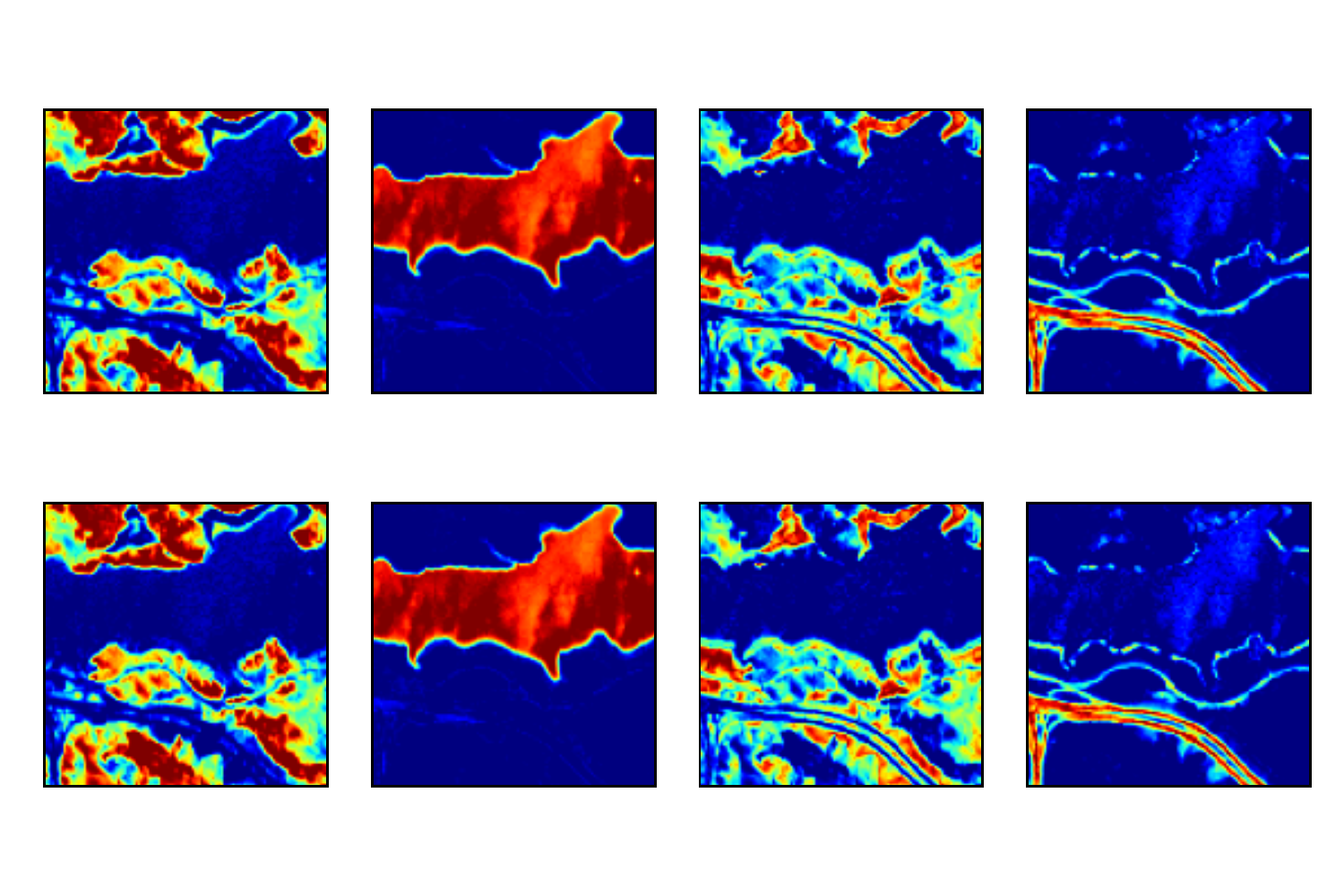}
    \caption{Top: Recovered. Bottom: True Abundance Maps for Jasper. The colormap is on scale from 0 (blue) to 1 (red).}
    \label{fig:jasper_abun}
\end{figure}
The metric values are reported in Table \ref{tab:err_lmm}, where our network performs better than others, for both the Samson and Jasper datasets. Although we obtain lower error values for MSE and SAD, the error performance of the neural architecture of End-Net \cite{ozkan2018endnet} is closest to ours.
\begin{table}[h]
    \centering
    \begin{tabular}{|c|cc|cc|} \hline
        Method &  \multicolumn{2}{c|}{Samson} &  \multicolumn{2}{c|}{Jasper} \\
         & MSE(A) & SAD(A) & MSE(A) & SAD(A) \\ \hline
         VCA &  3.86e-2 & 1.31e-1 & 1.60e-2 & 3.37e-1 \\
         MVSA & 6.29e-2 & 9.31e-2 &  3.80e-2 & 3.79e-1 \\
         Volmin & 7.48e-3 & 5.23e-2 & 9.47e-3 & 7.82e-2 \\
         DCAE & - & - & 9.58e-3 & 1.43e-1 \\
         EndNet & 1.50e-3 & 2.98e-2 & 6.33e-3 & 3.91e-2 \\
         Ours (S) & \textbf{9.41e-4} & \textbf{1.97e-2} & \textbf{2.97e-3} & \textbf{7.12e-3} \\ \hline
    \end{tabular}
    \caption{Metric comparison for different methods on the Samson and Jasper HSI datasets. '-' Not reported.}
    \label{tab:err_lmm}
\end{table}
For compute time analysis, we only compare against other neural approaches, as they utilize GPU parallelism. Results are shown in Table. \ref{tab:lmm_time}. Although DCAE is fast, its error metrics are worse (Table. \ref{tab:err_lmm}). End-net being the best in error comparison, we have $10 \times$ speedup over it. An important point is that the number of network parameters is explicitly equal to $2 \times K \times F$, irrespective of the image size, hence it is significantly low weight\footnote{ Further design guidelines can be found in Appendix.} compared to other deep nets. 
\begin{table}[h]
    \centering
    \begin{tabular}{|l|c|} \hline
        Method & Time (s) \\ \hline
        Auto-Enc \cite{palsson2018hyperspectral} & 1425 \\
        DCAE \cite{khajehrayeni2020hyperspectral} & 90 \\
        End-net \cite{ozkan2018endnet} & 855 \\
        Ours & \textbf{64} \\ \hline
    \end{tabular}
    \caption{Runtime on neural nets on Jasper Dataset}
    \label{tab:lmm_time}
\end{table}

\subsection{Case IV: Pareto Fronts}

A multi-objective optimization problem is formulated as:
\begin{align*}
    \underset{}{min} \quad F(x) &= (f_1(x),f_2(x),\ldots,f_k(x)) \numberthis \label{eq:multi}\\
    \text{s.t.} \quad x \in S &= \{ x \in \mathbb{R}^n | G(x)=(g_1(x), g_2(x),\ldots,g_m(x)\}
\end{align*}
in $n$ variables $(x_1,\ldots,x_n)$, $k$ objective functions $(f_1,\ldots,f_k)$, and $m$ constraint functions $(g_1,\ldots,g_m)$. The Pareto optimal solution $x_i$ satisfies the following conditions:
\begin{align*}
    &\nexists x_j: f_p(x_j) \leq f_p(x_i), \quad \textrm{for} \quad p=1,2,\ldots,k\\
    &\exists l: f_l(x_j) < f_l(x_i) \numberthis \label{eq:pareto}
\end{align*}
Fritz John Condition: Let the objective and constraint function in Eq. \ref{eq:multi} be continuously differentiable at a decision vector $x^* \in \mathcal{S}$. A necessary condition for $x^*$ to be Pareto optimal is that vectors must exists for $0 \leq \lambda \in \mathbb{R}^k$, $0 \leq \mu \in \mathbb{R}^m$ and $(\lambda, \mu) \neq (0,0)$ such that the following holds:
\begin{align*}
    \sum_{i=1}^k \lambda_i \nabla f_i(x^*) + \sum_{j=1}^m \mu_j \nabla g_j(x^*) = 0 \numberthis \label{eq:fjcond} \\
    \mu_jg_j(x^*) = 0, \forall j=1,\ldots,m
\end{align*}
\cite{gobbi2015analytical} defined the matrix $L$ as follows, comprising of the gradients of the functions and constraints. \begin{align*}
&L = \begin{bmatrix}
\nabla F & \nabla G \\
\mathbf{0} & G
\end{bmatrix} \quad [(n+m) \times (k+m)] \\
&\nabla F_{n \times k} = [\nabla f_1, \ldots, \nabla f_k]\\
&\nabla G_{n \times m} = [\nabla g_1, \ldots, \nabla g_m]\\
&G_{m \times m}=diag(g_1,\ldots,g_m)
\end{align*}
The matrix equivalent of Fritz John Condition for $x^*$ to be Pareto optimal, is to show the existence of $\lambda \in \mathbb{R}^{k+m}$ in Eq. \ref{eq:fjcond} such that the following holds true:
\begin{align}
    L \cdot \delta = 0 \quad \text{s.t.} \quad L=L(x^*),\delta \geq 0, \delta \neq 0 \label{eq:fjmatrix}
\end{align}
The non-trivial solution for Eq. \ref{eq:fjmatrix} is for the following to hold:
\begin{align}
    det(L^TL)=0 \label{eq:paropt}
\end{align}

We consider three cases of Pareto Fronts where the analytical solution is known. The key advantages of our approach compared to Genetic Algorithms \cite{gobbi2015analytical} and Scalarization \cite{ghane2015new} are:
\begin{itemize}
    \item Our network does not rely on the explicit form of the functions $F,G$. It can operate under discrete setting.
    \item Our network can operate on non-convex forms of $F,G$ compared to \cite{gobbi2015analytical}.
    \item Our network takes significantly less time compared to Genetic Algorithms.
    \item Our approach generates Pareto points uniformly with high density, where previous works \cite{ghane2015new} limit themselves to low density of points $\sim 40$.
\end{itemize}

\noindent \textbf{Case I} Gobbi \cite{gobbi2015analytical}: Jointly minimize
\begin{align*}
    f_1(x_1,x_2) = 1 - exp(-[(x_1-1/\sqrt(2))^2 + (x_2-1/\sqrt(2))^2]) \\
    f_2(x_1,x_2) = 1 - exp(-[(x_1+1/\sqrt(2))^2 + (x_2+1/\sqrt(2))^2])
\end{align*}
The analytical solution of the Pareto front is:
\begin{align*}
    f_1 = 1 + (f_2-1) \exp(-4+4 \sqrt{- \log (1-f_2)}) \\
    \text{under} \quad 0 \leq f_2 \leq 0.982, -\frac{1}{\sqrt{2}} \leq x_1,x_2 \leq \frac{1}{\sqrt{2}}
\end{align*}
This form is convex in $f_1,f_2$, wherein the approach proposed by Gobbi \cite{gobbi2015analytical} can find a solution.
\begin{figure}[h]
    \centering
	\includegraphics[width=0.4\linewidth]{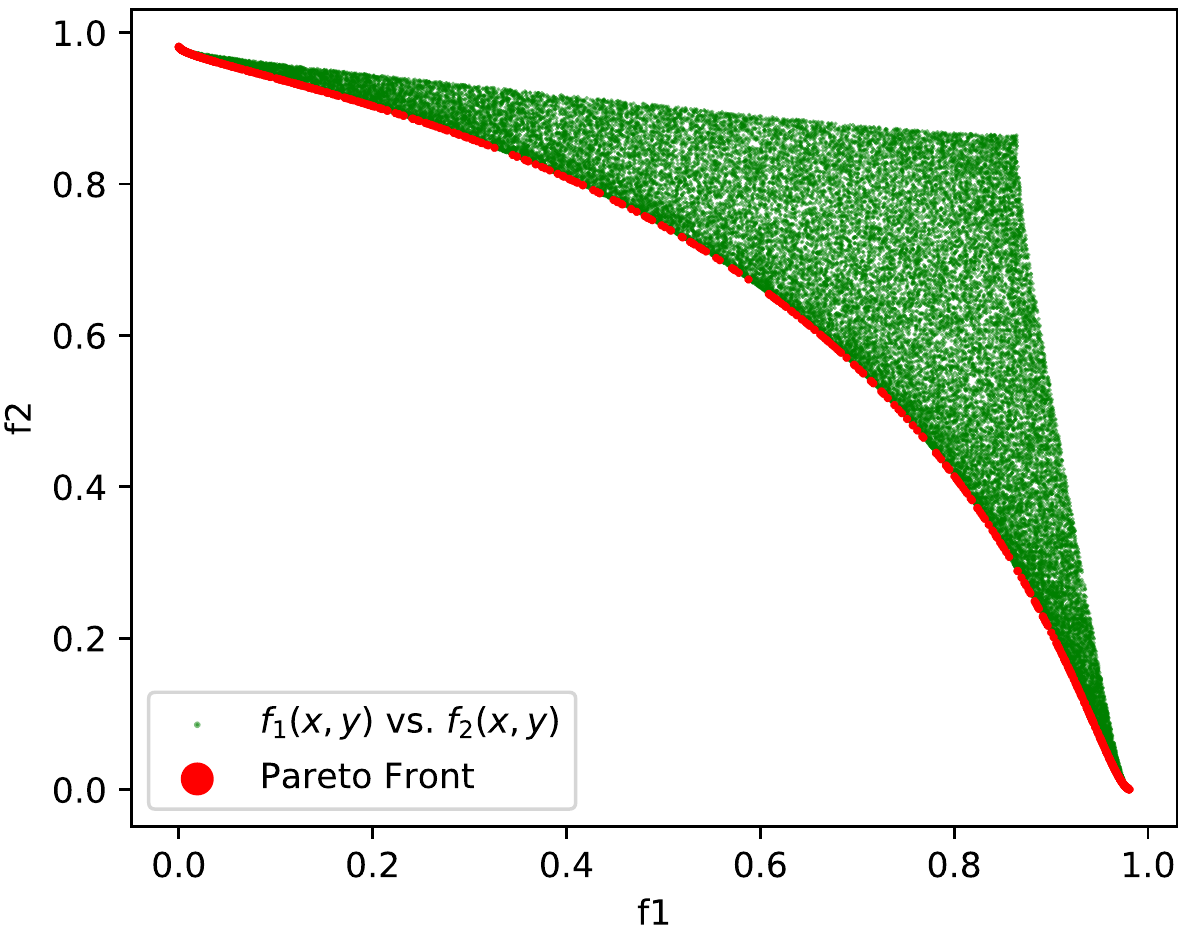}
    \caption{Pareto Front for Case I}
    \label{fig:pareto1}
\end{figure}

\noindent \textbf{Case II} Ghane \cite{ghane2015new}: Jointly minimize
\begin{align*}
    &f_1(x_1,x_2) = x_1 \\
    &f_2(x_1,x_2) = 1 + x_2^2 - x_1 - 0.1sin 3 \pi x_1\\
    &\text{s.t.} \quad 0 \leq x_1 \leq 1, -2 \leq x_2 \leq 2
\end{align*}
This form is non-convex in $f_2$, where the convex assumption of Gobbi \cite{gobbi2015analytical} does not hold. 
\begin{figure}[h]
    \centering
	\includegraphics[width=0.4\linewidth]{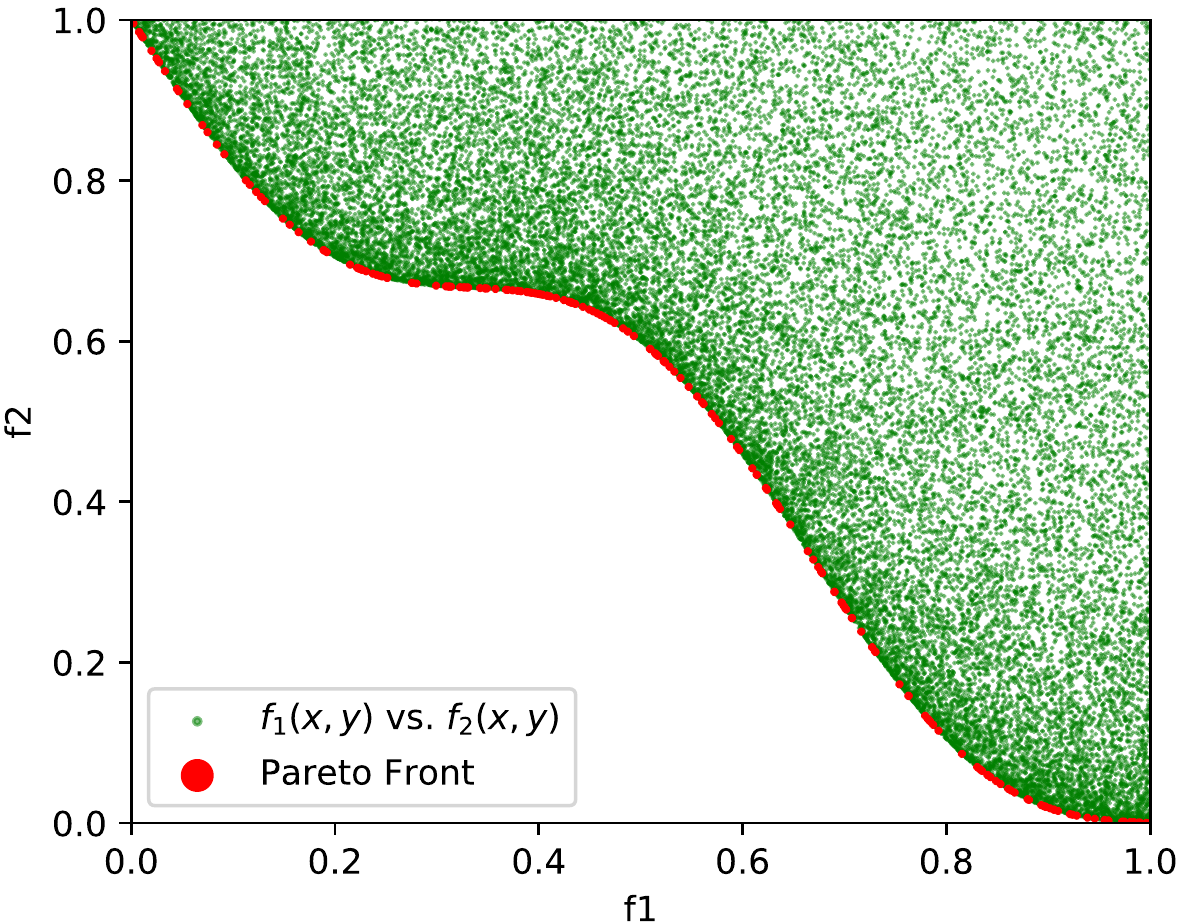}
    \caption{Pareto Front for Case II}
    \label{fig:pareto2}
\end{figure}

\noindent \textbf{Case III} Ghane \cite{ghane2015new}: Jointly minimize
\begin{align*}
    &f_1(x_1,x_2) = x_1 \\
    &f_2(x_1,x_2) = x_2\\
    &\text{s.t.} \quad \, g_1(x_1,x_2)= (x_1-0.5)^2 + (x_2-0.5)^2 \leq 0.5\\
    & g_2(x_1,x_2)= x_1^2 + x_2^2 - 1 - 0.1 \cos (16 \arctan (\frac{x_1}{x_2})) \geq 0\\
    &0 \leq x_1, x_2 \leq \pi
\end{align*}
This form is non-convex in $g_1,g_2$, where Gobbi \cite{gobbi2015analytical} exits. The Pareto surface for the above three forms are non-convex. Both our approach and Ghane \cite{ghane2015new} work on these forms, but our point density is significantly higher than those reported by them.
\begin{figure}[h]
    \centering
    \includegraphics[width=0.4\linewidth]{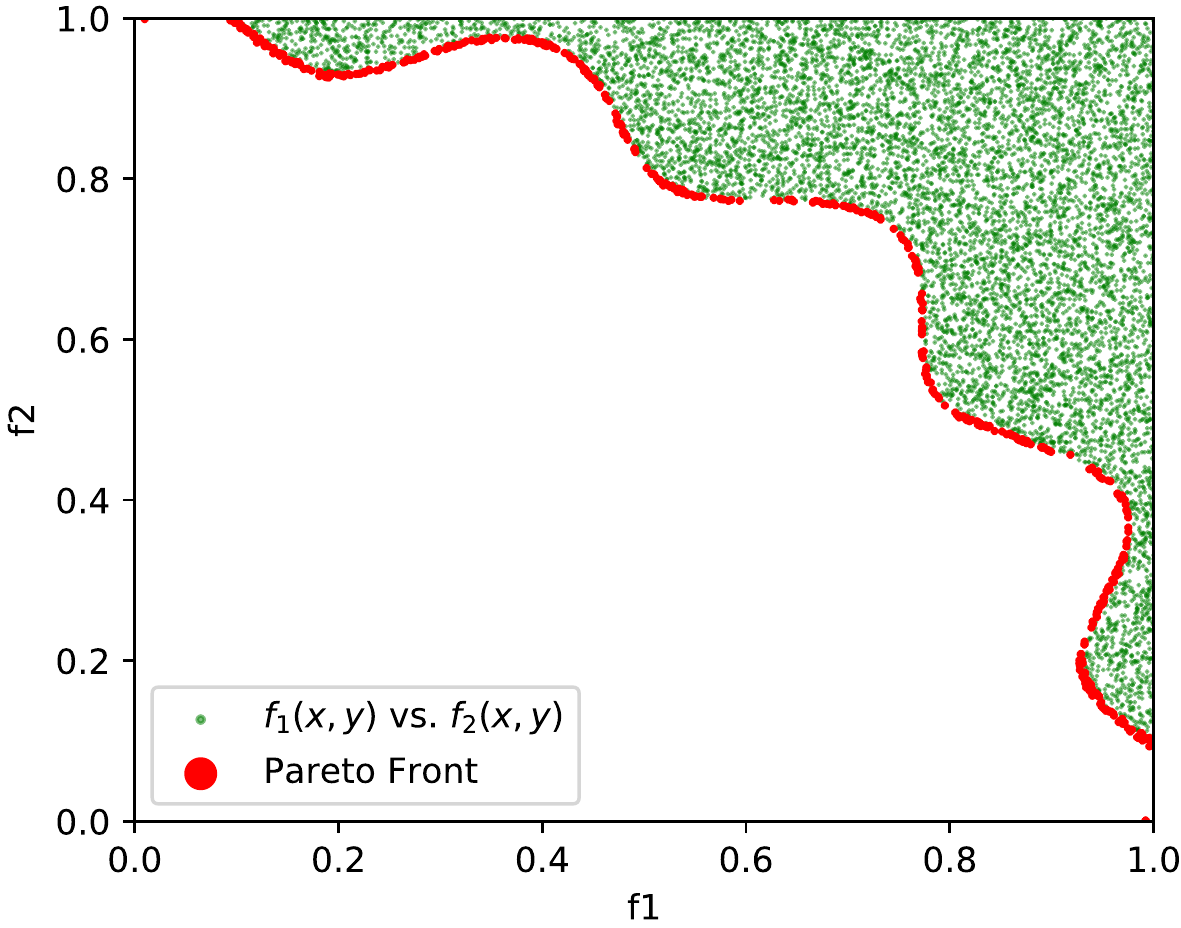}
	\caption{Pareto Front for Case III}
    \label{fig:pareto3}
\end{figure}
Our network loss satisfies the Fritz John necessary condition for Pareto optimality in Eq. \ref{eq:fjmatrix} and then uses a \textit{softmax} transformation to classify Pareto optimal points. All the experiments for $(w,d)=(8,4)$ and the input density of points is $(1000, 1000)$.

\begin{table}[h]
    \centering
    \begin{tabular}{|l|c|c|c|} \hline
        Method & \multicolumn{3}{c|}{Time (s)} \\
         & Form I & Form II & Form III \\ \hline
        Ours & 64 & 64 & 64 \\
        Gobbi \cite{gobbi2015analytical} & 123 & - & - \\
        Ghane \cite{ghane2015new} & 2.92 & 3.35 & 23.89 \\ \hline
    \end{tabular}
    \caption{Runtime comparison of the three methods. Notice that although, Ghane seems better in runtime, their method ran to detect just 33 points on the front, compared to the high density in our case shown in the figures.}
    \label{tab:my_label}
\end{table}

\section{Approximation Error}

Assuming the $L$ matrix From Eq. \ref{eq:fjmatrix} is square, we have
\begin{align*}
    det(L) = 0
\end{align*}
This is true for all the problems chosen in our numerical experiments. From Leibniz formula for determinants, we have:
\begin{align*}
    det(L) &= det\Big( \begin{bmatrix}
\nabla F & \nabla G \\
\mathbf{0} & G
\end{bmatrix} \Big) \\
&= det \Big( \begin{bmatrix}
\nabla F & 0 \\
0 & I
\end{bmatrix} \begin{bmatrix}
 I &  0 \\
0 & G
\end{bmatrix} \begin{bmatrix}
 I &  G \\
0 & I
\end{bmatrix}  \Big) \\
&= det(\nabla F) det(G) = 0
\end{align*}
Further assume that, 
\begin{align}
det(L(\tilde{x})) \leq \epsilon, \, \tilde{x} \neq x^* \label{eq:approx}
\end{align}
where $\epsilon > 0$. The Fritz John necessary condition in Eq. \ref{eq:fjcond} for weak Pareto optimality is:
\begin{align}
    det(L(x^*)) = 0 \label{eq:true}
\end{align}
Combining the assumption in Eq. \ref{eq:approx} and Eq. \ref{eq:true}, we have
\begin{align}
|det(L(\tilde{x}))-det(L(x^*))| \leq \epsilon
\end{align}
Notice that the manifold $h(x)=det(L(x))$. We further assume a low precision manifold $h(\tilde{x})$ such that:
\begin{align}
    \|h(\tilde{X}) - h(X^*)\|_2^{2} \leq \epsilon \label{eq:netopt}
\end{align}
When the network converges, Eq. \ref{eq:netopt} will hold for the network approximated $\tilde{h}(x)$. Here, $\tilde{x} \in \tilde{X} =\{x|\tilde{h}(x)=0\}$ and $X^*$ is the set of true optimal points such that $h(x^*) = 0 \forall x \in X^*$. Since we explicitly specify $\epsilon$ in our loss description, we know that the network generated solution is $\epsilon$ close to $h(\tilde{x})$ if:
\begin{align}
    \|h(\tilde{X}) - \tilde{h}(\tilde{X}) \|_{2}^{2} \leq C \epsilon, \quad 0 \leq C \leq 1 \label{eq:close}
\end{align}
The form in Eq. \ref{eq:close} implies that if we are able to find such a $C$, then we implicitly satisfy Eq. \ref{eq:netopt}. Hence, 
\begin{align}
    \|\tilde{h}(\tilde{X}) - h(X^*)\|_2^{2} \leq \epsilon
\end{align}

\section{Modeler Interpretability}

Motivated by the definition of interpretability and trust in model as stated in \cite{lipton2018mythos}, we state interpretability of our model from the viewpoint of a modeler. In all of the problems above, the approximate manifold $\tilde{h}$ is described by the user specified loss function. If a domain specific analytical solution is known $h(X^*)=0$, then the approximate network solution set $\tilde{h}(\tilde{X})=0$ can be verified by comparing $\tilde{X}$ and $X^*$. Additionally, a domain-specific modeler can also compare the approximate manifold $\tilde{h}$, at the penultimate layer from the output, against the true manifold $h$ known from the analytical form. This adds an extra layer of interpretation and increases trust in the network's working.

\section{Conclusion and Future Work}

A constrained neural optimization framework is presented for practical applications like non-linear solvers, implicit surfaces, hyperspectral unmixing and Pareto front detection. The network can operate with both limited data and functional forms to extract a lower dimensional solution manifold, described by the loss function. Numerical experiments against known analytical solutions show that the network approximates the implicit functions for both convex and non-convex problems. A modeler can confirm the results of our network by specifying a loss corresponding to an analytical solution known to them, thereby promoting interpretability. For HSI unmixing on real data, our network outperforms previously reported results both in terms of mean square error and spectral angle distance, with lower compute times. An important point for the unmixing problem is that, the network size (number of parameters) is explicitly dictated by $(2 \times \#$end-member $\times$ spectral length), irrespective of the image size.


\appendix

\section{End-Member Extraction}

In a Linear Mixture Model (LMM), each image pixel $y \in \mathbb{R}^F$ represents a data point and is a linear combination of a set of $K$ end-members $A=[a_1,\ldots,a_K] \in \mathbb{R}_+^{F \times K}$, with different fractions $b=[b_1,\ldots,b_K] \in \mathbb{R}_+^{K}$. In addition, $\gamma$ is an additive noise (assumed zero mean Gaussian) to account for sensor noise and illumination variation \etc. The LMM can thus be described as:
\begin{align} 
y = \sum_{k=1}^{K}a_kb_k + \gamma, \quad \text{s.t.} \, a_k,b_k \geq 0, \, \sum_{k=1}^Kb_k=1  \label{eq:mix}
\end{align}
Since we are interested in the minimum volume approaches, we seek to minimize the the volume of this convex cone to arrive at the solution.
\begin{align}
V(A) = \frac{|det(A)|}{K!}
\end{align}
The equivalent problem then becomes:
\begin{align}
\underset{A,B}{min} \, \|Y-AB\|_F^2 + \lambda det(A^TA)
\end{align}
Here $\lambda$ is a hyper-parameter that balances the trade-off between the regression loss and the minimum volume loss. 

\subsection{Network Architecture}

\begin{figure}[h]
	\centering
	\includegraphics[width=0.5\linewidth]{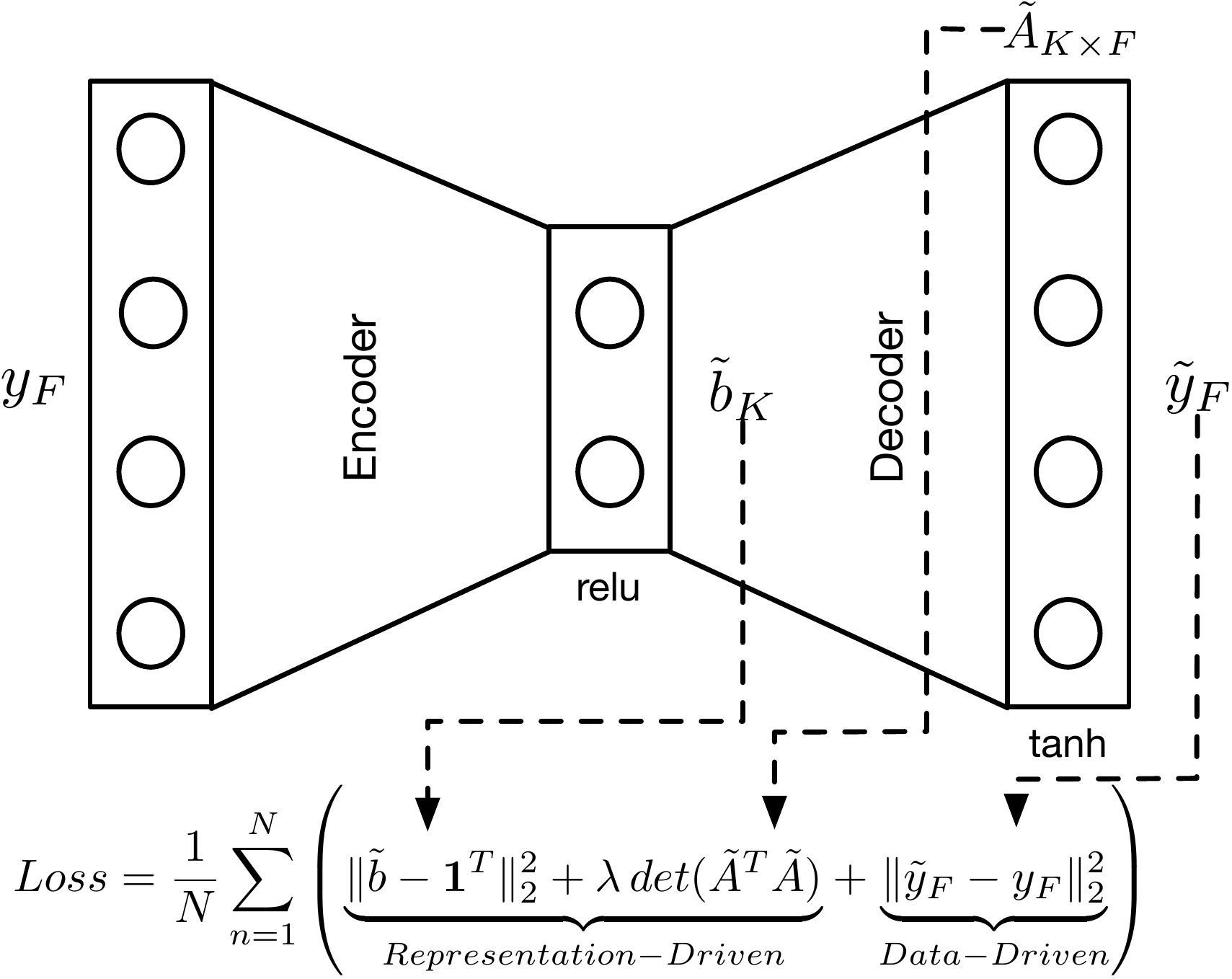}
	\caption{Architecture for End-Member Extraction with detailed notation for constraints driven loss}
	\label{fig:endarch}
\end{figure}

Fig. \ref{fig:endarch} shows the feed forward architecture tailored towards end-member extraction. It consists of two layers. The first one is of length $K$ and maps the input data to a $K$ dimensional plane, where a minimum volume simplex is constructed. The second (output) layer consists of $F$ nodes corresponding to the spectral length $F$. The weight matrix between layer one and two holds the invariant structure $A$, corresponding to the spectral responses of the end-members. The total number of parameters in our proposed auto-encoder is then $2FK$. Also note that there are no biases in any of the layers as the linear-mixture model for end-member extraction does require them. The first layer uses a \textit{relu} activation to enforce the criterion $b_k \geq 0$. The last layer has \textit{tanh} activation is used for a smooth reconstruction of the input signal.

\subsection{Auto-Encoder Representation}

For the auto-encoder to represent the input $y_{F} \in V$, the encoder and decoder weights, $\hat{A}_{F,K}$ and $\tilde{A}_{K,F}$ must satisfy the following bi-orthogonality property:

\begin{equation}
\hat{A}\tilde{A}^{T} = I_{K}
\end{equation}

In other words, if $\hat{A}$ is the dual of $\tilde{A}$ then $\hat{A}^{T}\tilde{A} \in V_{K} \subset V$ such that:  
\begin{equation}
\tilde{y}_{F} = f_{dec}(f_{enc}(y_{F}\hat{A}^{T})\tilde{A}) \approx y_{F}\hat{A}^{T}\tilde{A}
\end{equation}

This is trivially true if the encoder and decoder activation functions $f_{enc}$ and $f_{dec}$ are linear equipped with a Frobenius norm (data-driven loss). However, for linear activation functions we do not have $s = y_{F}\hat{A}^{T}$, where s lie on a simplex (abundances). For this specific choice of representation, the simplex and volume minimization criteria are additionally enforced as representation driven loss with encoder and decoder activation functions: $f_{enc} = relu$ and $f_{dec}=tanh$.

\begin{figure*}[h]
	\centering
	\includegraphics[width=\linewidth]{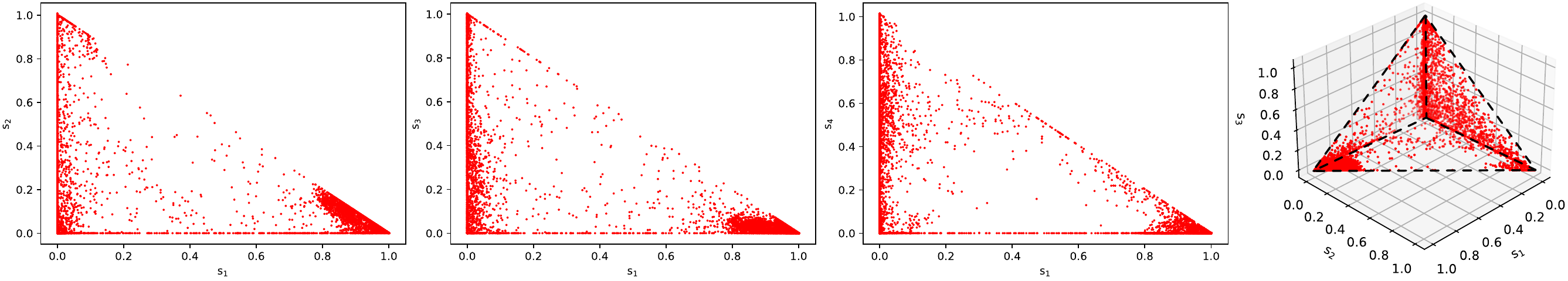}
	\caption{Recovered simplex for Jasper}
	\label{fig:jasper_tet}
\end{figure*}

\subsection{Further Results}

\begin{figure}[ht]
	\centering
	\includegraphics[width=0.5\linewidth]{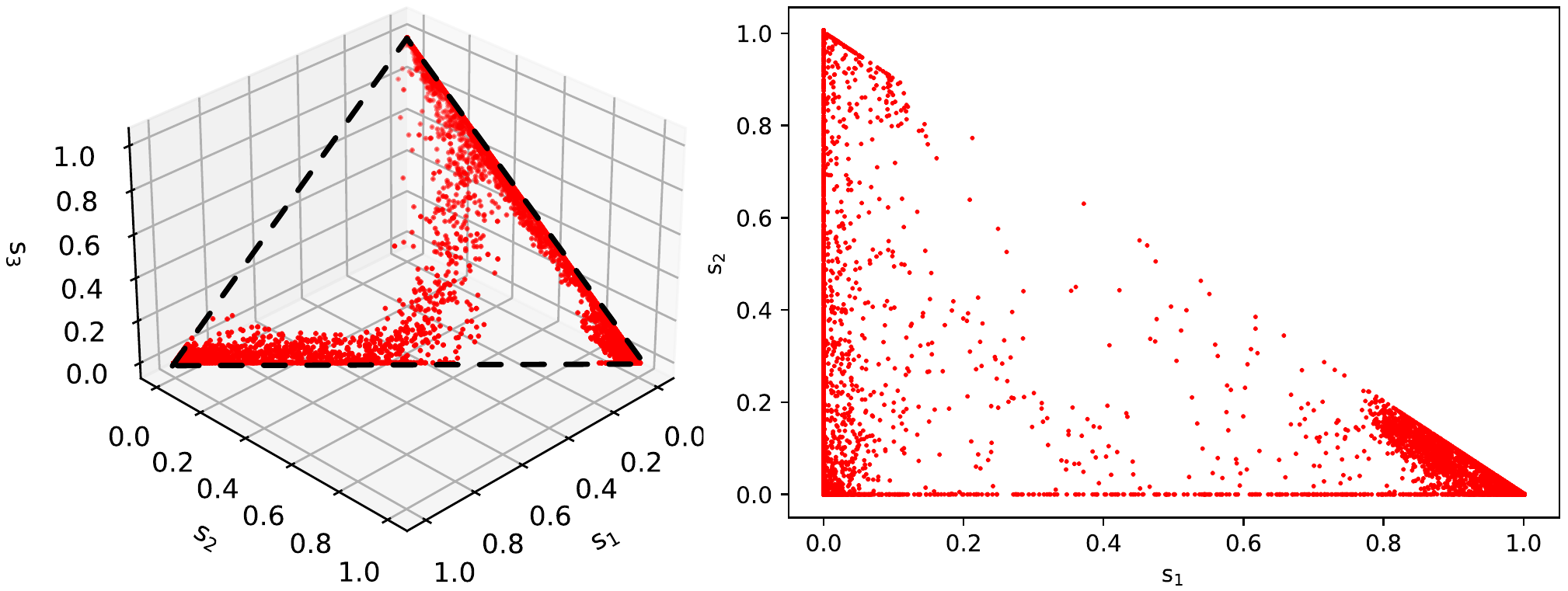}
	\caption{Recovered simplex for Samson}
	\label{fig:samson_tet}
\end{figure}

The recovered simplices for both the Samson and the Jasper datasets are also shown in \ref{fig:jasper_tet} and  Fig. \ref{fig:samson_tet}. Note that under a good recovery, the sides of the quad be of equal length. In other words, projection onto a 2D space, each view should be as close right angled triangle as possible.

Fig. \ref{fig:jasper_abun} shows the true and recovered abundance maps for the Jasper dataset. Having attained $MSE \sim 10^{-4}$ using our proposed network, the two maps are visually identical in the colormap scale of $[0-1]$.

\begin{figure}[h]
	\centering
	\includegraphics[width=\linewidth]{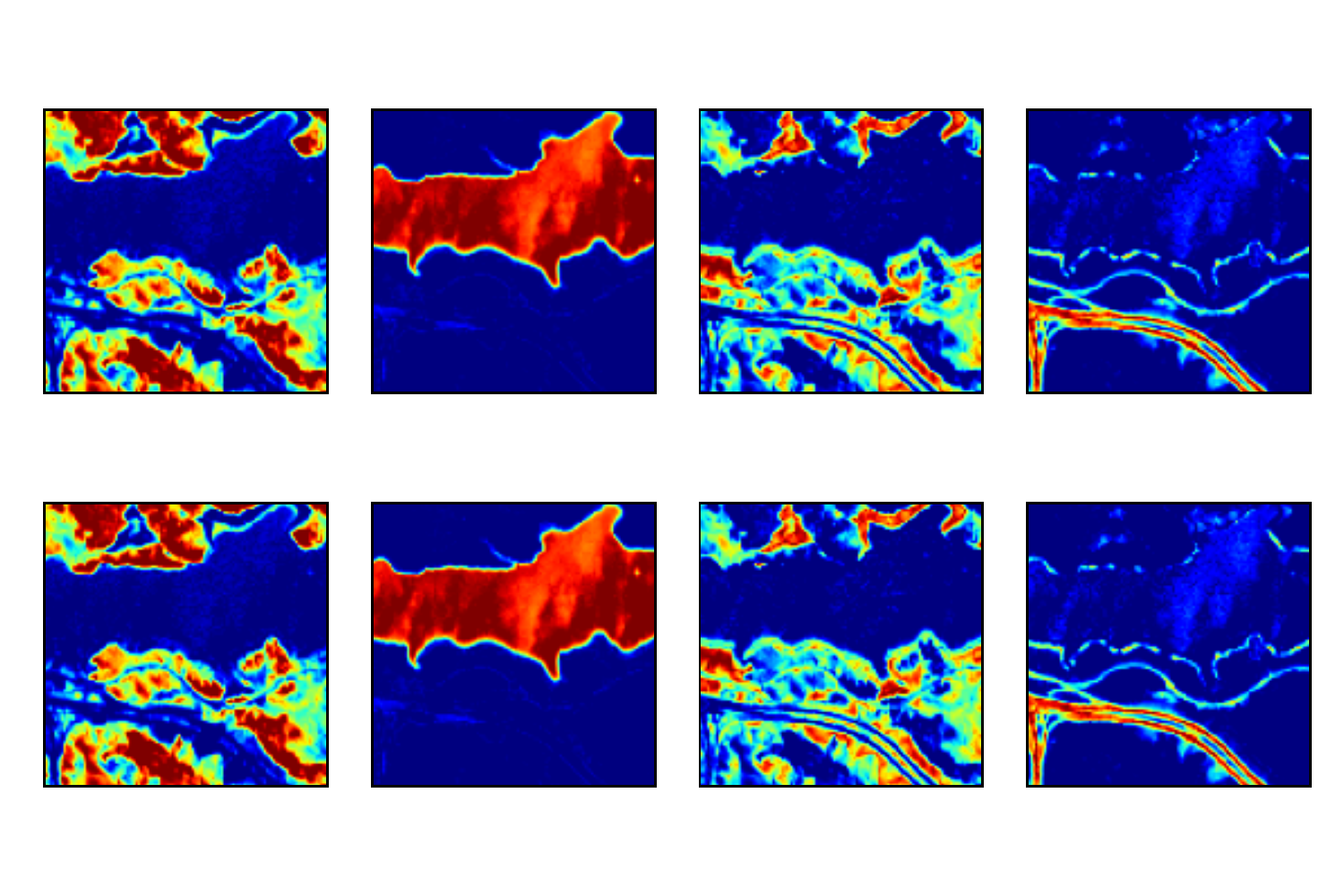}
	\caption{Top: Recovered. Bottom: True Abundance Maps for Jasper. The colormap is on scale from 0(blue) to 1(red).}
	\label{fig:jasper_abun}
\end{figure}

\end{document}